\documentclass[conference]{IEEEtran}
\IEEEoverridecommandlockouts
\usepackage{cite}
\usepackage{amsmath,amssymb,amsfonts}
\usepackage{algorithmic}
\usepackage{graphicx}
\usepackage{textcomp}
\usepackage{xcolor}
\usepackage{multirow}
\def\BibTeX{{\rm B\kern-.05em{\sc i\kern-.025em b}\kern-.08em
    T\kern-.1667em\lower.7ex\hbox{E}\kern-.125emX}}
\begin{document}

\title{An End-to-End Framework For Universal Lesion Detection With Missing Annotations\\
}

\author{\IEEEauthorblockN{Xiaoyu Bai, Yong Xia}
\IEEEauthorblockA{\textit{School of Computer Science and Engineering, Northwestern Polytechnical University, Xi'an, China}\\
Email: bai.aa1234241@mail.nwpu.edu.cn, yxia@nwpu.edu.cn}
}

\maketitle

\begin{abstract}
Fully annotated large-scale medical image datasets are highly valuable. However, because labeling medical images is tedious and requires specialized knowledge, the large-scale datasets available often have missing annotation issues. For instance, DeepLesion, a large-scale CT image dataset with labels for  various kinds of lesions, is reported to have a missing annotation rate of 50\%. Directly training a lesion detector on it would suffer from false negative supervision caused by unannotated lesions. To address this issue, previous works have used sophisticated multi-stage strategies to switch between lesion mining and detector training. In this work, we present a novel end-to-end framework for mining unlabeled lesions  while simultaneously training the detector. Our framework follows the teacher-student paradigm. In each iteration, the teacher model infers the input data and creates a set of predictions. High-confidence predictions are combined with partially-labeled ground truth for training the student model. On the DeepLesion dataset, using the original partially labeled training set, our model can outperform all other more complicated methods and surpass the previous best method by 2.3\% on average sensitivity and 2.7\% on average precision, achieving state-of-the-art universal lesion detection results.
\end{abstract}


\section{Introduction}
\begin{figure}
    \centering
    \includegraphics[scale=0.4]{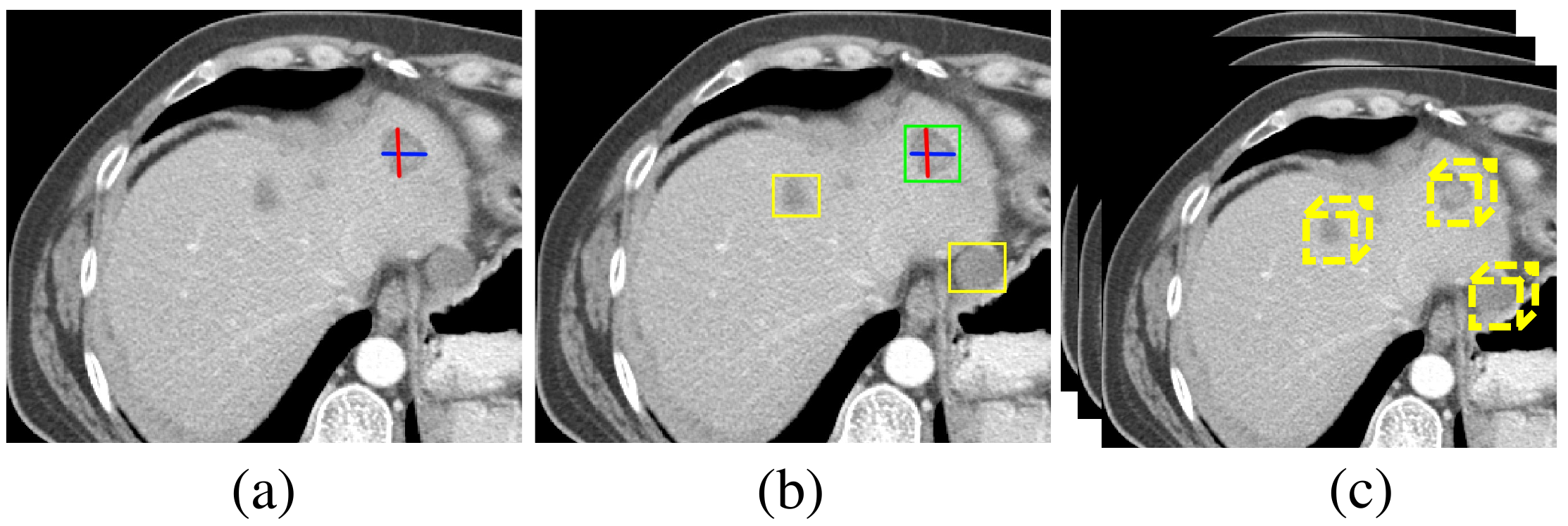}
    \caption{Annotations in DeepLesion dataset. (a) The original RECIST  annotation. (b) The RECIST generated bounding box is indicated in green, and two missing annotations are shown in yellow. (c) There is no lesion depth information, therefore the 3D bounding boxes are unavailable. }
    \label{fig:1}
\end{figure}

Universal lesion detectors (ULDs) \cite{tang2019uldor,yan20183d}, which aim to discover various types of lesions in multiple organs on radiological images, fit the daily clinical need of general radiologists and thus have significant clinical values.
Due to the success of deep learning in computer vision, most ULDs are deep convolutional neural networks (DCNNs), which require a large-scale annotated image dataset to train.
Unlike labeling natural images, annotating medical images is costly and requires extensive domain knowledge \cite{baid2021rsna}.  The various appearances of lesions on medical images make it even more challenging to annotate all lesions thoroughly. For instance, the large-scale CT image dataset with RECIST annotations \cite{eisenhauer2009new} for various kinds of lesions, namely DeepLesion, is reported to have 50\% lesions unlabeled (see Fig. \ref{fig:1}). 
If we train a ULD on DeepLesion, those unlabeled lesions will be probably misinterpreted as the background, resulting in false negative supervision signals and degrading performance in lesion detection. While ULD has been extensively studied in \cite{tang2019uldor,li2019mvp,tao2019improving,yan2019mulan,zhang2020revisiting}, this issue was only recognized by a few recent works \cite{wang2019semi,cai2020lesion,yan2020learning,lyu2021segmentation}.
The key to tackling this missing annotation issue is to filter out the false negative samples generated by unlabelled lesions. Based on this idea, current methods can be  categorized into two groups: sampling based and pseudo-label based methods. Sampling based methods aim to find reliable background regions that are not unlabelled objects. For instance, 
overlap-based hard sampling \cite{shrivastava2016training} only samples negative box proposals with an overlap larger than 0.1 to any existing ground truth annotations. 
The soft sampling method \cite{wu2018soft} re-weights negative regions based on their overlaps and the center distances with annotated instances. While these strategies can reduce the number of false negative samples, they ignore numerous informative areas that include valuable true negatives. 
\begin{figure*}
    \centering
    \includegraphics[scale=0.6]{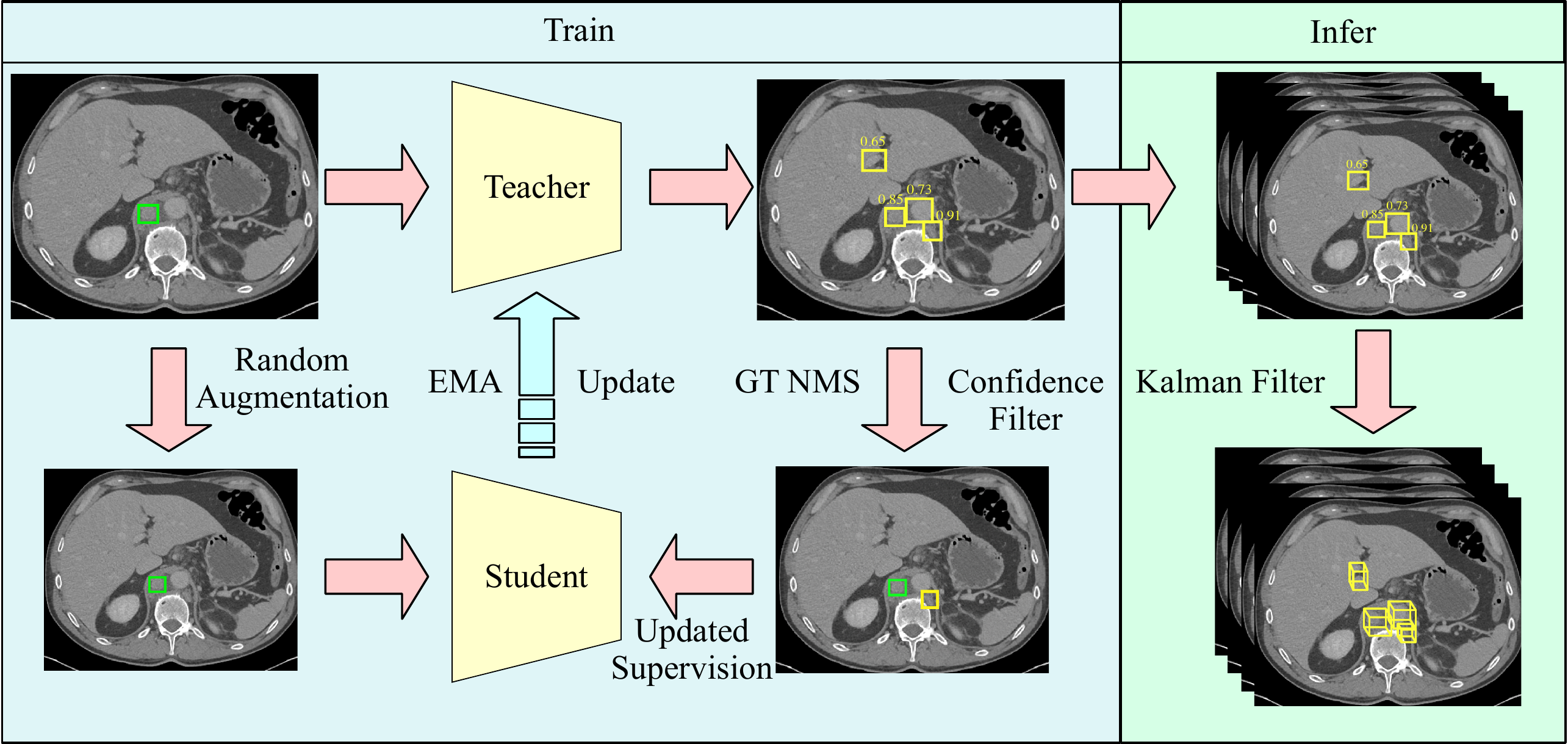}
    \caption{Diagram of the proposed model lesion mining model. }
    \label{fig:framework}
\end{figure*}
To handle the disadvantages of the sampling based method, pseudo-label based methods consider directly mining unlabelled lesions and using them to guide the model training. Wang et al. \cite{wang2019semi} advocate using the continuity of multi-slice axial CT images to mine unlabeled lesions. Yan et al. \cite{yan2020learning} use the knowledge gained from several fully-labeled single-type lesion datasets to assist the ULD training. Cai et al. \cite{cai2020lesion} introduce the lesion-harvester framework, which combines a lesion proposal generator and a lesion proposal classifier to iteratively mine unlabeled lesions and hard negatives for retraining. The previous state-of-the-art method \cite{lyu2021segmentation} adopts the co-teaching idea and proposes a 
dual-task model to mutually check the suspicious unlabeled lesions mined by each task.
Although these previous pseudo-label based methods improve the overall detection performance, they use complex multi-stage procedures to transition between lesion mining and detector training,  which limits their applicability to larger datasets \cite{tarvainen2017mean}. Furthermore, all these methods require the detectors to be trained to convergence on each stage, while the mined lesions are only updated once.  We argue these multi-stage approaches are not optimal, both on the training efficiency and the performance.
A better approach may be gradually mining the lesions and bootstrapping the model performance in a more fine-grained process, i.e., in batches, so that the false negative samples can be corrected timely during training.

Based on this concept, we present a model for processing lesion mining on each mini-batch. The performance gained by this change is significant. For instance, on the DeepLesion dataset, in \cite{cai2020lesion} it takes three days to run six iterations for mining the unlabelled lesion, while the average precision (AP) performance only increases from 43.0\% to 43.6\%. The previous state-of-the-art method \cite{lyu2021segmentation} achieves the AP performance to 47.5\%, but it also relies on extra data processing and fine-tuning steps. In contrast, our method can boost the AP performance to 50.2\% by a single end-to-end training process in eight hours.

Specifically, our method adopts the teacher-student framework to simultaneously mine unlabeled lesions and train the detector. For each mini-batch, the teacher branch infers the input data and creates a set of predictions.  The high-confidence predictions are considered suspicious lesions and are combined with partially-labeled ground truth to train the student branch. The teacher branch maintains an exponential moving average (EMA) of the student's parameters and is used as the final output model. The main contributions are summarized as follows: (1)We propose a novel end-to-end framework to address the missing annotation problem in the universal lesion detection task. (2) We systematically compared different policies for dealing with the mined suspicious regions.
(3) The proposed framework is highly efficient and achieves the state-of-the-art lesion detection performance on the fully-annotated test set of the DeepLesion dataset.

\section{Method}

Fig. \ref{fig:framework} illustrates the overview of our lesion mining and detection model. It follows the teacher-student paradigm  \cite{tarvainen2017mean}, consisting of two detection branches with the same structure. The parameters of the student branch are updated by gradient back-propagation under the supervision of the partially labeled ground truth and the  mined suspicious lesions from the teacher branch. The teacher branch maintains an exponential moving average (EMA) of the student's parameters. We use Faster R-CNN \cite{ren2015faster} as our lesion detector framework.

Given the partially annotated dataset $D_p = \{x_i,y_i^l,y_i^u\}_{i=1}^N$, where  $x_i$ represents the image and $y_i^l$ is the available lesion annotations, $y_i^u$ denotes the missing annotations that are not accessible. For the generation of suspecious lesions, at each mini-batch, the teacher branch first infers the batch inputs $\{x_i\}_{i=1}^{N_b}$ to give a set of predictions $\{p_i^{\rm{T}} \}_{i=1}^{N_b}$. Next, the predictions $\{p_i^{\rm{T}} \}_{i=1}^{N_b}$ are compared with the available ground truth $\{y_i^l\}_{i=1}^{N_b}$ by their intersection over union (IoU) values, we then remove the predictions which have IoU larger than a threshold with any ground truth. For simplicity, we use $P^{\rm{T}}$, $Y^l$ and $Y^u$ to denote the batch predictions  $\{p_i^{\rm{T}} \}_{i=1}^{N_b}$, available ground truth $\{y^l\}_{i=1}^{N_b}$  and unlabeled ground truth $\{y^u\}_{i=1}^{N_b}$. The remaining predictions  can be written as $P^{\rm{T}} - P^{\rm{T}} \cap Y^l$. This process can be easily implemented using Non-Maximum Suppression (NMS) operation on the predictions and ground truth, we name it as GT NMS operation. Finally, we filter out the predictions with confidence scores lower than threshold $\tau $. The remaining predictions are represented as $P^m$ and will be served as mined lesions to feed the student branch.

For the training of the student branch, consider the student branch's predictions  $P^{\rm{S}}=\{p_i^{\rm{S}} \}_{i=1}^{N_b}$ for the resized batch input images. The detection loss can be summarized as a combination of position classification loss and box regression loss:
\begin{equation}
  \mathcal{L} = \mathcal{L}_{cls}(P^{\rm{S}},Y) + \mathcal{L}_{reg}(P^{\rm{S}},Y),
\end{equation}
where $Y = Y^l+Y^u$ is the ground truth.
Since the box regression loss $\mathcal{L}_{reg}$ is only computed on each annotated object, it is not influenced by the false negatives issue. However, for the classification loss $\mathcal{L}_{cls}$, it needs to assign each position to one object or background class. Suppose there are $M$ positions, each position represents a prior (anchor) box $b_i$ and is assigned to one lesion in $Y$ or background label. For simplicity, we regard all lesions in $Y$ as foreground label.  The number of foreground boxes and background boxes are $M^f$ and $M^b$. The ideal classification loss is:
\begin{equation}
    \mathcal{L}_{cls} = \dfrac{M^f}{M} \sum_{i=1}^{M^f}H(c_i^f,+1) +  \dfrac{M^b}{M} \sum_{i=1}^{M^b}H(c_i^b,-1),
\end{equation}
where $c_i^f$ and $c_i^b$ represent the model's output foreground and background probability at each position, $H(t,y)$ indicates the loss  between the model's outputs  $t$  and the ground truth $y$.

We can further separate $M^f$ into $M^l$ and $M^u$, corresponding to $Y^l$ and $Y^u$. Since the $Y^u$ are unavailable, the $M^u$ will be falsely assigned to background label. Thus the real loss is a biased version:
\begin{equation}
    \mathcal{L}_{cls}' = \dfrac{M^l}{M} \sum_{i=1}^{M^l}H(c_i^f,+1) +  \dfrac{M^b+M^u}{M} \sum_{i=1}^{M^b+M^u}H(c_i^b,-1).
\end{equation}
The error is:
\begin{equation}
    \mathcal{L}_{cls}' - \mathcal{L}_{cls} = -\dfrac{M^u}{M} \sum_{i=1}^{M^u}H(c_i^f,+1) +  \dfrac{M^u}{M} \sum_{i=1}^{M^u}H(c_i^b,-1).
    \label{error}
\end{equation}
If the teacher mined lesions $P^m$ contains any unlabeled ground truth in $Y^u$, we can use it to reduce the error (\ref{error}). Considering that $P^m$ may contain false positive results, most methods choose to ignore the mined lesions.
This policy could only reduce the second term in the error. On the contrary, benefitting from the flexibility of our framework's online mining ability and the EMA teacher's better capability, we can gradually add the teacher's most confident predictions as pseudo labels to train the student branch. The final classification loss for student branch is:
\begin{align}
\begin{split}
   & \hat{\mathcal{L}_{cls}} = \dfrac{M^l+M^m}{M} \sum_{i=1}^{M^l+M^m}H(c_i^f,+1) \\
   & +\dfrac{M^b+M^u-M^m}{M} \sum_{i=1}^{M^b+M^u-M^m}H(c_i^b,-1).
\end{split}
\end{align}

The mining frequency is flexible. Here, we mine suspicious lesions on every mini-batch. The total mining times for each image is equal to the training epochs and is decided by the performance on the fully annotated validation set.

We use the teacher branch as our final model because its EMA update strategy has a model ensemble effect \cite{tarvainen2017mean} which can give better results. It is vital for the ensemble model to have diverse components. However, the commonly used learning rate decay policy will make the student model converge to similar parameters. Therefore, we use a constant learning rate without decay in our training. For inference, We use the same Kalman filter as lesion harvester \cite{cai2020lesion} for tracing the 2d predictions on each slice to the 3d lesion bounding boxes.
\begin{figure}[tb]
    \centering
    \includegraphics[scale=0.16]{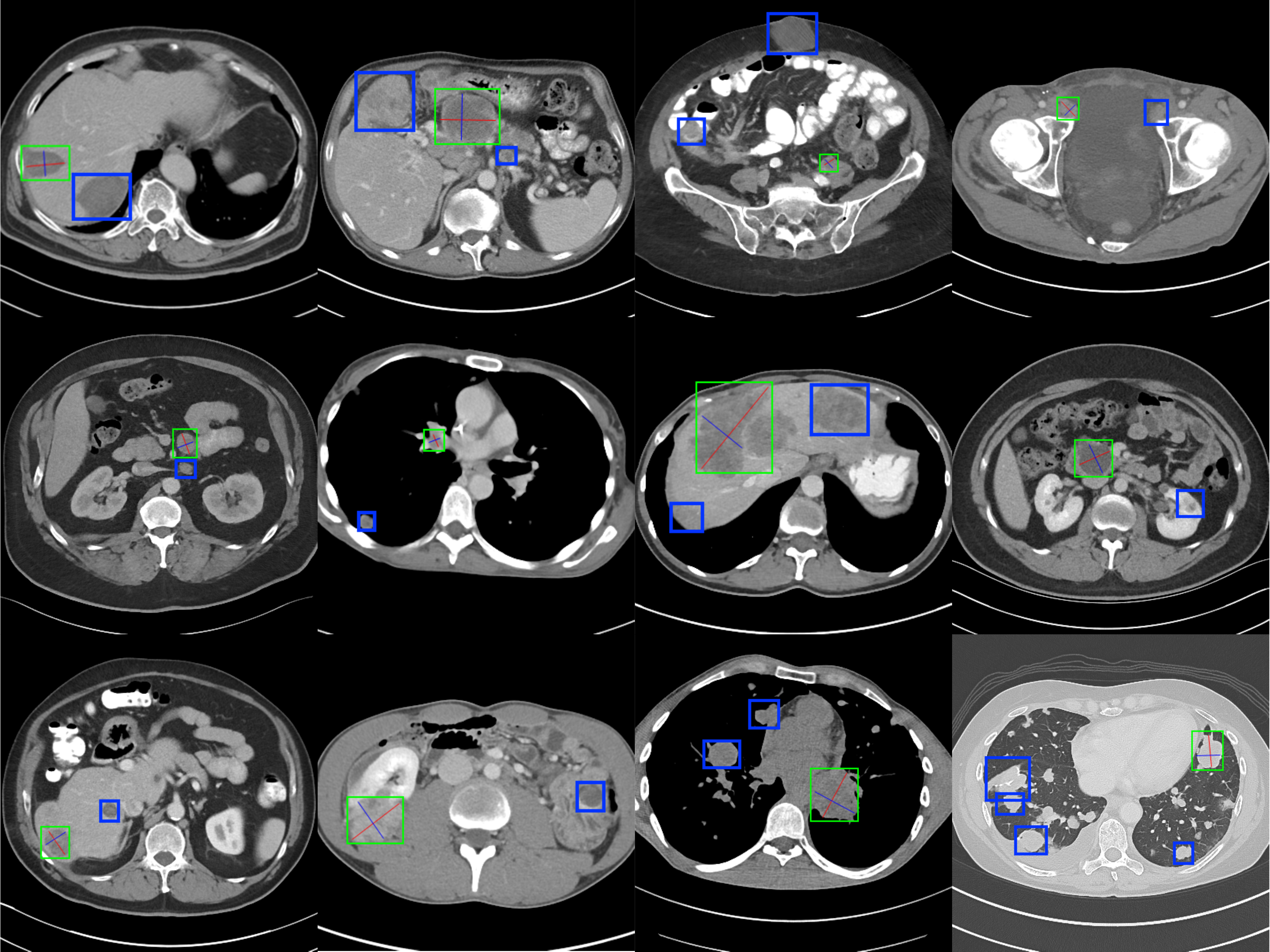}
    \caption{An illustration of the mined suspicious lesions on the DeepLesion training set. The original RECISTs produced annotations are represented by the green bounding boxes. The mined lesions with output confidence greater than 0.9 are indicated by the blue bounding boxes. }
    \label{fig:mined}
\end{figure}

\section{Experiments}
\subsection{Dataset and Metrics}
 DeepLesion \cite{yan2018DeepLesion} is the largest CT dataset for universal lesion detection at the moment, but it contains roughly 50\% unlabeled lesions \cite{yan20183d}. The original test set also has the missing annotation issue and could cause unreliable evaluation results. Cai et al. \cite{cai2020lesion} fully annotated and released 1915 of the DeepLesion subvolumes. 844 subvolumes of them are selected from the original training set, in which 744 subvolumes are used for additional training, 100 subvolumes for validation. The other 1,071 subvolumes are used for the test. We also adopt this setting. Specifically, we use the same 100 fully annotated subvolumes to validate our model, the rest of the original training set to train our model. The 1071 fully annotated subvolumes are used for evaluation.

 Similarly to previous studies, we report the average precision (AP) and  the detection sensitivities at different false positive (FP) rates  (0.125 to 8) per volume using the P3D IoU evaluation metric \cite{cai2020lesion}.

 \begin{table*}
\centering
\caption{Average precision (AP) and sensitivity at various FPs per image on the fully annotated testing set of DeepLesion, the official annotated lesions, mined suspicious lesions by different methods and mined hard negative samples are denote as R, P+ and P- respectively.}
\setlength{\tabcolsep}{3mm}
\begin{tabular}{@{}l|ccc|cccccccc|c@{}}
\hline
 \multicolumn{1}{c|}{\multirow{2}{*}{Method}}&  &  &  & \multicolumn{8}{c|}{Sensitivity (\%) at different FPs per subvolume} &  \\ \cline{5-12}
 & \multirow{-2}{*}{R} & \multirow{-2}{*}{P+} & \multirow{-2}{*}{P-} & 0.125 & 0.25 & 0.5 & 1 & 2 & 4 & 8 & AVG & \multirow{-2}{*}{AP (\%)} \\ \hline
 Mask R-CNN \cite{he2017mask} & { \checkmark} &  &  & 13.64 & 19.92 & 27.48 & 36.73 & 47.01 & 57.41 & 66.38 & 38.36 & 41.4 \\
Yan et al. \cite{yan2019mulan} &  \checkmark &  &  & 11.43 & 18.69 & 26.98 & 38.99 & 50.15 & 60.38 & 69.71 & 39.47 & 41.8 \\
Cai et al. \cite{cai2020lesion} & { \checkmark} &  &  & 11.92 & 18.42 & 27.54 & 38.91 & 50.15 & \textbf{60.76} & \textbf{69.82} & 39.64 & 43.0 \\

Lyu et al. \cite{lyu2021segmentation} & { \checkmark} &  &  & 14.84 & 20.08 & 27.93 & 37.14 & 47.59 & 57.65 & 66.47 & 38.81 & 42.1 \\
Ours & { \checkmark} &  &  & \textbf{18.63} & \textbf{25.45} & \textbf{32.78} & \textbf{40.43} & \textbf{49.00} & 57.51 & 65.67 & \textbf{41.32} & \textbf{45.0} \\
\hline
Wang et al. \cite{wang2019semi} & \checkmark & \checkmark &  & 15.88 & 20.98 & 29.41 & 38.36 & 47.89 & 58.14 & 67.89 & 39.79 & 43.2 \\
Cai et al. \cite{cai2020lesion} & \checkmark & \checkmark &  & 13.41 & 19.16 & 27.34 & 37.54 & 49.33 & 60.52 & 70.18 & 39.64 & 43.6 \\
Lyu et al. \cite{lyu2021segmentation} & \checkmark & \checkmark &  & 20.01 & 25.65 & 33.44 & 42.81 & 52.39 & 60.69 & 68.79 & 43.39 & 47.5 \\
Ours & \checkmark & \checkmark &  & \textbf{21.17} & \textbf{28.13} & \textbf{36.10} & \textbf{45.73} & \textbf{54.98} & \textbf{63.47} & \textbf{70.67} & \textbf{45.75} & \textbf{50.2} \\ \hline
Cai et al. \cite{cai2020lesion} & \checkmark & \checkmark & \checkmark & 19.86 & 27.11 & 36.21 & 46.82 & 56.89 & 66.82 & \textbf{74.73} & 46.92 & 51.9 \\
Lyu et al. \cite{lyu2021segmentation} & \checkmark & \checkmark & \checkmark & \textbf{30.89} & \textbf{37.28} & \textbf{43.94} & 51.05 & 57.41 & 63.68 & 69.81 & 50.58 & 54.7 \\
Ours & \checkmark & \checkmark & \checkmark & 27.63 & 35.87 & 42.37 & \textbf{51.68} & \textbf{58.93} & \textbf{66.87} & 74.65 & \textbf{51.14} & \textbf{55.3} \\\hline
\end{tabular}

\label{tab:comparison}
\end{table*}
\subsection{Implementation Details}
Our model is built on Faster R-CNN \cite{ren2015faster} using MMDetection toolbox \cite{mmdetection}.
We employ the ImageNet pretrained truncated DenseNet-121 \cite{huang2017densely}  as the backbone, and each input contains 9 contiguous CT slices, as in earlier work \cite{cai2020lesion,lyu2021segmentation}. the CT slices are interpolated to 2 mm  in the z-axis and 0.8 mm in the xy-plane. We rescale the 12-bit CT intensity range to  [0,255] using a single windowing (-1024–3071 HU).
We use random resize as the data augmentation method and the resize ratio for student model is [0.8,1.2]. The IoU threshold for GT NMS operation is 0.7 and the confidence filter threshold for mined lesion is set to $\tau=0.9$. The momentum value for EMA update is set to 0.999.
We adopt stochastic gradient descent (SGD) optimizer to train the model for 80000 iterations with the learning rate $lr = 0.02$. We do not use any learning rate decay schedule. The final model is chosen based on the performance on the validation set.
It takes about 8 hours to train our model for 80000 iterations on 4 RTX-2080Ti GPUs. 
\subsection{Main Results}

 Table \ref{tab:comparison} compares the performance of the proposed method and earlier state-of-the-art ULD methods on the fully annotated DeepLesion test set.
 MULAN \cite{yan2019mulan} is one of the best-performing methods on the original partially labeled test set. Wang et al. \cite{wang2019semi}, Cai et al. \cite{cai2020lesion}, and Lyn et al. \cite{lyu2021segmentation} are three recent methods considered to mined the unlabeled lesions. In Lyn et al.'s \cite{lyu2021segmentation} paper, the performance of the Mask-RCNN method is also reported. We also list it in the table. 
 We first show the results trained only using the original annotations. As we can see, the EMA ensembled teacher branch achieves the overall best performance. In particular, it outperforms the second best model by 1.67\% of the average sensitivity and 2.0\% average precision, which demonstrates the useful ensemble effect of the EMA update policy. 
 
 For the lesion mining results, our method surpasses all other methods by a large margin. specifically, it outperforms the current state-of-the-art method \cite{lyu2021segmentation} by 2.3\% of the average sensitivity and 2.7\% average precision, proving that our end-to-end method has better lesion mining capacity. We show some lesion mining results in Fig. \ref{fig:mined}. The green bounding boxes represent the original annotations. The mined lesions are shown in blue. In total, we mined about 8000 suspicious lesions on the DeepLesion training set.
 
 Finally, for the completeness of comparative experiments, same to the method used in \cite{lyu2021segmentation}, we also  apply our trained detector on the fully annotated training subvolumes and select the hard negative samples for retraining. Compared with \cite{lyu2021segmentation},  our method can find more lesions  and show the best overall metric results.
 
\subsection{Ablation Studies}
 We start by verifying that the confidence filter's threshold value is appropriate. We have done this by training our model for a total of 20000 iterations with various threshold settings. Table \ref{tab:conf} displays the result. We report the average value of the  sensitivitity at different FPs (AS) and average precision (AP). When the threshold is set to 0.9, the best results are obtained. In general, the higher the threshold, the more true positives it will reject, whereas the lower the threshold, the more false positives it will produce. When the  threshold value is set to 0.80, the performance rapidly degrades due to the addition of too many false positive samples for training the student branch.
 
Next, we tested different strategies for using the mined lesions. The mined lesions are ignored by the previous approach \cite{lyu2021segmentation}. We tested this strategy and set the ignore range as $[0.9,1)$ corresponding to the confidence filter's threshold. The results are shown in Table \ref{tab:stra}. Both of these strategies bring substantial performance gain and adding the mined lesions to ground truth can give best results.
 
 \begin{table}
\centering
\caption{Ablation study on different confidence filter threshold.}
\begin{tabular}{c|c|c}
\hline
$\tau$ value  & AS (\%) & AP (\%) \\
\hline
0.95 & 42.93 & 46.90\\
0.90 & 43.72 & 47.91 \\
0.85 & 41.49 & 45.04 \\
0.80 & 30.99 & 31.25 \\

\hline
\end{tabular}
\label{tab:conf}
\end{table}

 \begin{table}
\centering
\caption{Comparison of different  strategies for using the mined lesions.}
\begin{tabular}{@{}c|c|c@{}}        
\hline
Strategy  & AS (\%) & AP (\%) \\
\hline
Adding & 45.75 & 50.21\\
Ignoring & 43.63 & 48.96 \\
\hline
\end{tabular}
\label{tab:stra}
\end{table}

 \begin{table}
\centering
\caption{Ablation study on the learning rate schedule.}
\begin{tabular}{@{}c|c|c@{}}        
\hline
LR Schedule  & AS (\%) & AP (\%) \\
\hline
with LR decay & 42.21 & 46.10\\
w/o LR decay  & 44.10 & 48.57 \\
\hline
\end{tabular}
\label{tab:lr}
\end{table}
Learning rate decay is a commonly used technique for training neural networks. However, using the learning rate decay will harm the teacher's performance in our scenario since our model mimics the ensemble model's behavior. To demonstrate this, we first trained our model for 20000 iterations. On top of that, we trained another 20000 iterations with or without using learning rate decay.  Table \ref{tab:lr} displays the results. It is clear that the model trained without learning rate decay outperforms the other.

\section{Conclusion}
 In this paper, we present a novel end-to-end approach to dealing with missing annotations in the universal lesion detection task. We employ the teacher-student framework to mine unlabeled lesions and train the detector concurrently. 
Specifically, in each iteration, the teacher model infers the input data and creates a set of predictions.  The high-confidence predictions are considered suspicious lesions and combined with partially-labeled ground truth to train the student model on the fly. The teacher model maintains an exponential moving average (EMA) of the student model and is used as the final output model. 
We tested two policies for dealing with the mined suspicious regions: ignoring them or adding them as mined pseudo ground truth. We discovered that learning rate decay would degrade the teacher's performance and should not be employed. Instead, using a fixed learning rate can give us promising results.
We evaluated our model against the state-of-the-art methods on the DeepLesion dataset. By using the original partially labeled training set, our model can outperform the previous best method by 2.3\% on average sensitivity and 2.7\% on average precision. 
Our method can also be used in other lesion detection problems where complete annotations is hard to acquire.

\bibliographystyle{IEEEtran}
\bibliography{IEEEtran}
\end{document}